\documentclass[10pt,twocolumn,letterpaper]{article}

\usepackage{cvpr}              
\usepackage[accsupp]{axessibility}  
\usepackage{times}

\usepackage{algorithm}
\usepackage{algorithmic}

\usepackage{amsmath} 
\usepackage{amsthm}
\usepackage{thm-restate}
\usepackage{amsfonts}
\usepackage{physics}
\usepackage{subcaption}
\usepackage{booktabs}
\usepackage{multirow}
\usepackage{makecell}

\newtheorem{theorem}{Theorem}

\definecolor{cvprblue}{rgb}{0.21,0.49,0.74}
\usepackage[pagebackref,breaklinks,colorlinks,allcolors=cvprblue]{hyperref}

\def\paperID{12502} 
\def\confName{CVPR}
\def\confYear{2025}

\title{Adaptive Non-uniform Timestep Sampling for \\ Accelerating Diffusion Model Training}

\author{Myunsoo Kim$^{1*}$ \quad Donghyeon Ki$^{1*}$ \quad Seong-Woong Shim$^{1}$ \quad Byung-Jun Lee$^{1,2}$\\
$^1$Korea University \quad $^2$Gauss Labs Inc.\\
{\tt\small \{m970326, peop1e1n, ssw030830, byungjunlee\}@korea.ac.kr}}

\begin{document}
\maketitle
\def\thefootnote{*}\footnotetext{These authors contributed equally.}\def\thefootnote{\arabic{footnote}}
\begin{abstract}
As a highly expressive generative model, diffusion models have demonstrated exceptional success across various domains, including image generation, natural language processing, and combinatorial optimization. However, as data distributions grow more complex, training these models to convergence becomes increasingly computationally intensive. While diffusion models are typically trained using uniform timestep sampling, our research shows that the variance in stochastic gradients varies significantly across timesteps, with high-variance timesteps becoming bottlenecks that hinder faster convergence. To address this issue, we introduce a non-uniform timestep sampling method that prioritizes these more critical timesteps. Our method tracks the impact of gradient updates on the objective for each timestep, adaptively selecting those most likely to minimize the objective effectively. Experimental results demonstrate that this approach not only accelerates the training process, but also leads to improved performance at convergence. Furthermore, our method shows robust performance across various datasets, scheduling strategies, and diffusion architectures, outperforming previously proposed timestep sampling and weighting heuristics that lack this degree of robustness.
\vspace{-0.5cm}
\end{abstract}    
\section{Introduction}
\label{Introduction}
In recent years, diffusion models have demonstrated their high expressive power and achieved significant success across a wide range of domains, including image \citep{dhariwal2021diffusion, rombach2022high, nichol2021glide}, video \citep{khachatryan2023text2video, zheng2024open}, text \citep{li2022diffusion, austin2021structured, he2022diffusionbert}, audio generation \citep{schneider2023archisound}, combinatorial optimization \citep{sun2023difusco, sanokowski2024diffusion}, reinforcement learning \citep{ajay2022conditional, janner2022planning, wang2022diffusion}, and more. However, with the increasing complexity of target data distributions, achieving convergence in diffusion model training has become increasingly computationally demanding. For example, training images with Stable-Diffusion-2.0 \citep{rombach2022high} requires 24,000 A100 GPU hours, while Open-Sora \citep{zheng2024open} requires 48,000 H800 GPU hours for video training. These high computational costs not only pose significant barriers to advancing generative AI applications but also have negative environmental consequences.

While numerous studies have explored ways to accelerate diffusion model training, our research specifically targets modifications to the uniform training process of diffusion models across diffusion timesteps. It is well recognized that non-uniform training schemes over timesteps can enhance training speed. For example, several heuristics have been proposed~\citep{choi2022perception,hang2023efficient,wang2024closer,karras2022elucidating}.

However, the underlying reasons for the success of these methods remain largely unexplored, and they are often regarded as heuristics that may have potential limitations in robustness across diverse experimental conditions.

\begin{figure}[t]
    \centering
    \includegraphics[width=1.0\linewidth]{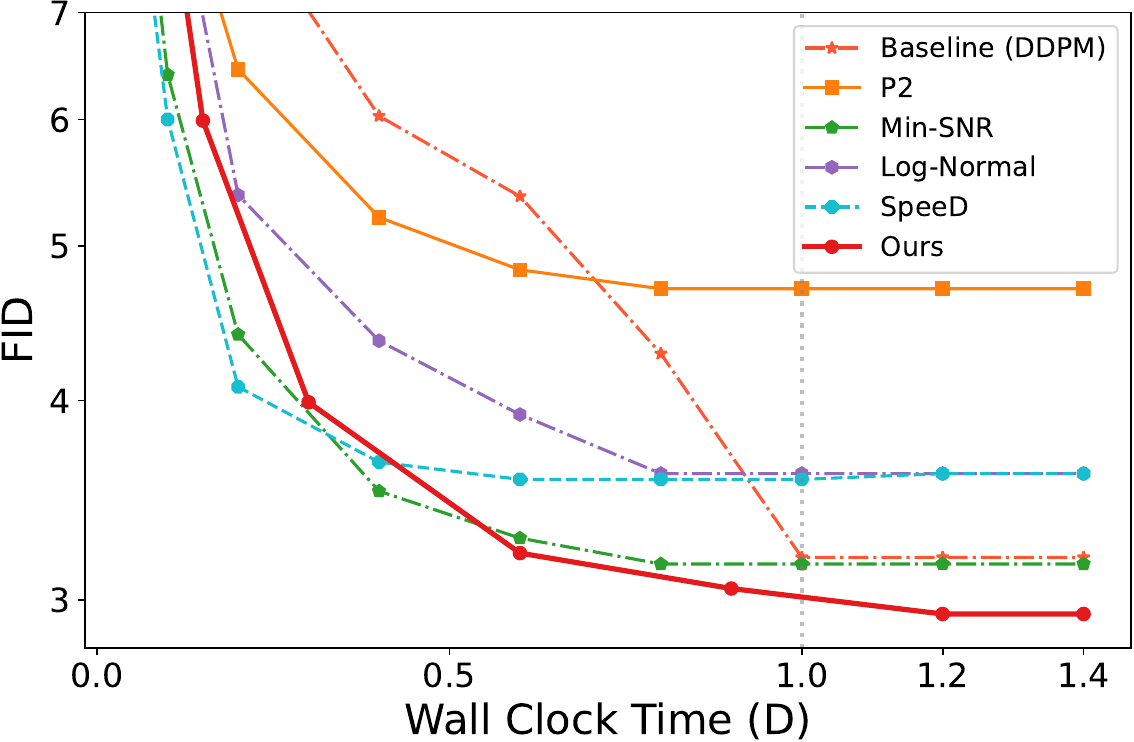}
    \caption{Comparison of FID scores of our approach and other acceleration methods against relative wall clock time, where 1D represents the time it takes for the baseline to converge. Although our learning method is initially slower than the heuristics due to its learning-based nature, it converges to a point with better optimality within 1D, and achieves a significantly lower FID score by 1.2D.}
    \label{fig:main_wallclocktime}
    \vspace{-0.3cm}
\end{figure}

Through a series of experiments, we observed significant variation in the stochastic gradient variance across different diffusion training timesteps. 

We hypothesize that this variability plays a key role in the imbalanced training observed with uniform sampling and may explain the effectiveness of previously proposed heuristics in improving diffusion model training. However, due to the strong interdependence of gradients across different timesteps, we found that this insight alone does not directly lead to optimal acceleration in diffusion training.

To this end, we propose a more direct approach by developing an efficient algorithm designed to sample timesteps that are estimated to yield the greatest reduction in the objective. This algorithm approximates the impact of gradient updates for each timestep on the variational lower bound and increases the sampling frequency of specific timesteps that require further optimization.

This enables the diffusion training process to converge significantly faster than previous heuristics across various settings.
To summarize, our three main contributions are as follows:
\begin{enumerate}
    \item Through an analysis of gradient variance in diffusion model training, we offer an explanation for why a non-uniform training process across timesteps can lead to faster convergence.
    \item Unlike previous heuristic-based diffusion acceleration methods, we propose a learning-based approach that adaptively samples timesteps by approximately minimizing the variational bound on the negative log-likelihood.
    \item We conduct experiments across various image datasets, noise scheduling strategies and diffusion architectures to demonstrate the robustness of our method.
\end{enumerate}
\section{Related Works}
\label{related works}

\paragraph{Non-uniform training of diffusion models} Several studies have explored timestep-wise non-uniform training of diffusion models to accelerate training. These approaches can be categorized into two main groups:
\begin{itemize}
    \item \textbf{1) Weighting methods:} These methods apply hand-designed weightings to the loss at each timestep. For example, P2~\citep{choi2022perception} assigns greater weights to timesteps containing perceptually rich content and lesser weights to those involving imperceptible details.~Similarly, Min-SNR~\citep{hang2023efficient} adjusts loss weights based on clamped signal-to-noise ratios.
\item \textbf{2) Sampling methods:} These approaches use hand-designed samplers to select timesteps in a non-uniform manner.~For instance, Log-Normal \citep{karras2022elucidating} samples timesteps from a log-normal distribution. SpeeD \citep{wang2024closer} modifies sampling probabilities to downweight timesteps with lower losses and prioritize those with rapidly changing process increments.
\end{itemize}
While these heuristics have proven effective in accelerating diffusion model training, their direct relationship to the underlying objective remains unclear, raising concerns about their applicability in new, potentially more effective settings that may emerge in the future.
\vspace{-0.3cm}
\paragraph{Learning to optimize (L2O)} L2O is a field in machine learning focused on learning optimization algorithm. It typically involves a meta-training set composed of multiple training and validation datasets. Using this meta-training set, L2O learns to update parameters based on gradients to minimize validation set errors, either through supervised learning~\citep{andrychowicz2016learning, wichrowska2017learned} or reinforcement learning~\citep{li2016learning, bello2017neural}. Although our objective similarly focuses on learning to optimize, we address a more challenging task by not assuming meta-training set for transfer learning. Instead, we focus on learning to optimize in an online manner, without the benefit of prior training data.

\begin{figure*}[htb!]
    \centering
    \includegraphics[width=1.0\linewidth]{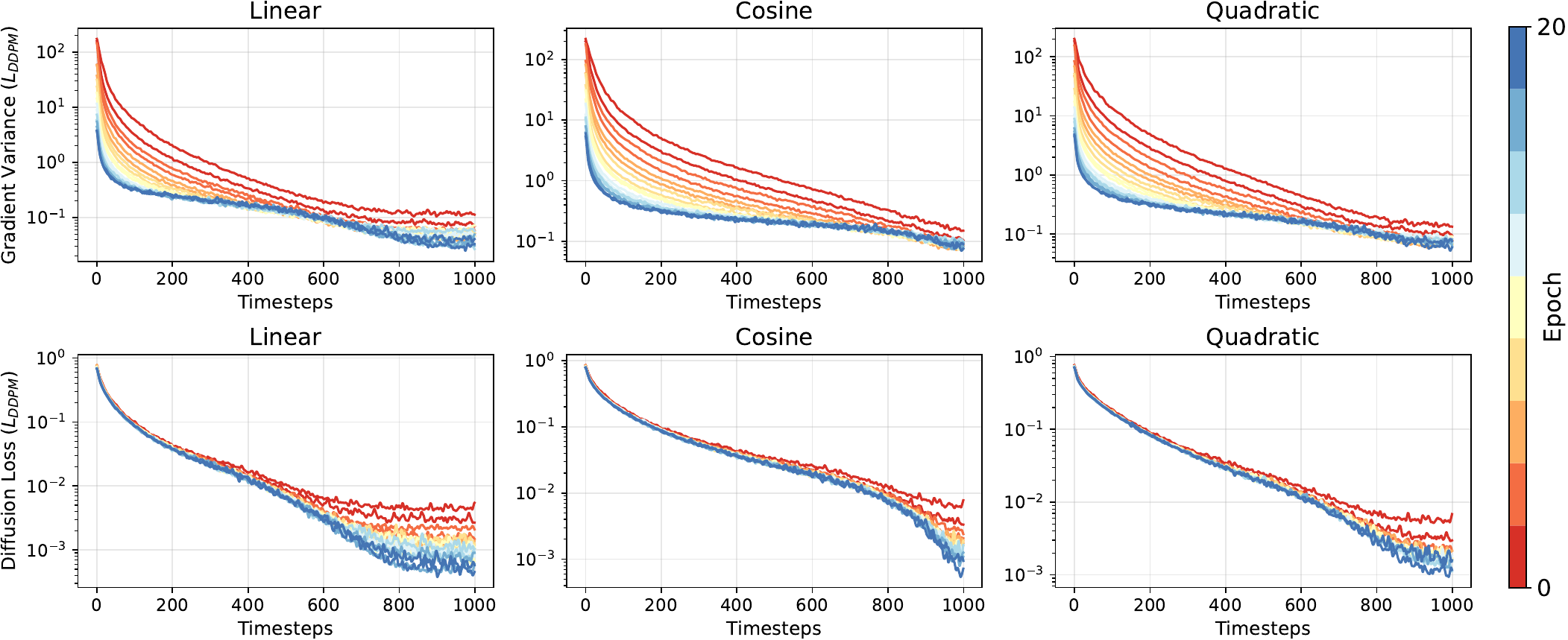}
    \caption{Comparison of gradient variance (top) and diffusion loss (bottom) over timesteps for three different noise schedules: linear, cosine, and quadratic. The color bar on the right indicates the progression of epochs, with red representing the early epochs and blue representing the later epochs. The results are based on training DDPM~\citep{ho2020denoising} with $\mathcal{L}_{\text{DDPM}}$ for 20 epochs.}
    \label{fig:gradient_variance_kl_divergence}
\end{figure*}
\section{Background: Diffusion Models}
Diffusion model \citep{sohl2015deep,ho2020denoising} consists of two key processes: a predefined forward noising process and a reverse denoising process, which is learned through variational inference to match the forward process. 
\vspace{-0.3cm}
\paragraph{Forward process} Starting with data $x_0 \sim q(x_0)$, the forward process operates as a Markov-Gaussian process that incrementally introduces noise, generating a sequence of noisy latent variables $\{x_1, x_2, \dots, x_T\}$:
\begin{align}
    q(x_t|x_{t-1}) &:= \mathcal{N}\left(x_t; \sqrt{1-\beta_t} x_{t-1}, \beta_t I\right), 
\end{align}
where $\beta_t$ is a noise schedule. Latent variables can be equivalently sampled by $x_t := \sqrt{\bar{\alpha}_t}x_0 + \sqrt{1- \bar{\alpha}_t} \epsilon$, $\epsilon \sim \mathcal{N}(0, I)$, where $\bar{\alpha}_t := \prod^t_{s=1} (1-\alpha_s)$ and $\alpha_t := 1 -\beta_t$. 
\paragraph{Reverse process} Diffusion models generate data $x_0$ from $x_T$ through the learned reverse process $p_\theta(x_{t-1}|x_t)$, gradually denoising the latent variables from a Gaussian noise:
\begin{align}
    p_\theta(x_{t-1}|x_t) &:= \mathcal{N}\left(x_{t-1}; \mu_\theta(x_t,t), \sigma_t^2 I \right), \\
\mu_\theta(x_t,t)&:=\frac{1}{\sqrt{\alpha_t}}\left(x_t-\frac{\beta_t}{\sqrt{1-\bar{\alpha}_t}}\epsilon_\theta(x_t, t)\right)
\end{align}
where $\sigma^2_t$ is a variance of the denoising process, and $\mu_\theta$ is parameterized as proposed by \citet{ho2020denoising} for further simplification. $\epsilon_\theta$ is a deep neural network with parameter $\theta$ that predicts the noise vector $\epsilon$ based on $x_t$ and $t$.
\paragraph{Training} The training of a diffusion model is achieved by optimizing the variational lower bound on the log-likelihood, which is expressed as a sum of KL-divergence terms comparing Gaussian distributions and can be reformulated in terms of the normalized noise:
\begin{align}
    \mathcal{L}_{\text{VLB}}&(\theta) := \mathbb{E}_{x_0}\left[\sum_{t} D_{\text{KL}}\left(q(x_{t-1}|x_t,x_0) \Vert p_\theta (x_{t-1}|x_t)\right)\right] \nonumber \\
    & = \mathbb{E}_{x_0, \epsilon, t}\left[\frac{\beta_t^2}{2\sigma^2_t \alpha_t(1-\bar{\alpha}_t)}\lVert \epsilon - \epsilon_\theta (x_t,t) \rVert^2\right],
\end{align}
where $t \sim \mathcal{U}\{1,T\}$. A standard way to train a diffusion model is to ignore the constant terms~\citep{ho2020denoising}, resulting in an objective of:
\begin{align}
\label{eq:L_simple}
    \mathcal{L}_{\text{DDPM}}(\theta) := \mathbb{E}_{x_0, \epsilon, t}\left[\lVert \epsilon - \epsilon_\theta ( \sqrt{\bar{\alpha}_t}x_0 + \sqrt{1- \bar{\alpha}_t} \epsilon,t) \rVert^2\right].
\end{align}
\vspace{-0.3cm}

\begin{figure*}[ht]
    \centering
    \includegraphics[width=1.0\linewidth]{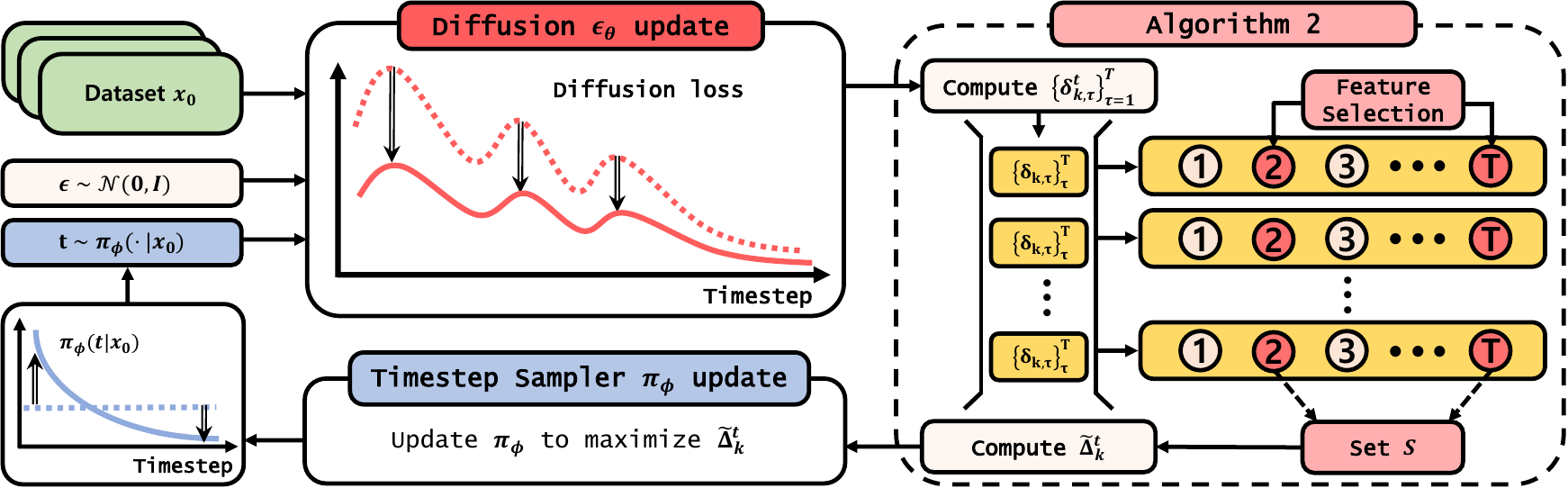}
    \caption{\textbf{Adaptive Non-Uniform Timestep Sampling.} 
    Visualization of the overall architecture of our algorithm. Following an update from $\theta_k$ to $\theta_{k+1}$, we compute the set $\{\delta^t_{k,\tau}\}_{\tau=1}^T$ based on $\theta_k$ and $\theta_{k+1}$, and add it to a queue. A feature selection method is then applied to identify $|S|$ timesteps that best explains ${\Delta}^t_k$ from this queue.  Then a timestep sampler $\pi_\phi$ is trained to maximize $\tilde{\Delta}^t_k$, which samples timestep
    s $t$ that is expected to achieve the largest reduction in the diffusion loss. This process continues iteratively to  minimize the diffusion loss at the chosen timesteps.}
    \vspace{-0.2cm}
    \label{fig:overall_algorithm}
\end{figure*}

\section{Demystifying Non-Uniform Diffusion Model Training Across Timesteps}

\label{section: Diffusion Model Training with Adam Optimizer}

In this section, we offer a hypothesis explaining why non-uniform training strategies may offer advantages over the uniform training approach used by \citet{ho2020denoising}. For simplicity, we begin by assuming an independent parameterization across timesteps, expressed as:
\begin{equation}
    \epsilon_\theta(x_t, t)=\epsilon_{\theta_t}(x_t),\quad \forall t,
\end{equation}
with a independent set of parameters $\theta=\{\theta_t\}_{t=1}^T$ for each timestep. This parameterization allows us to have independent optimization objective per timestep, i.e., $\mathcal{L}_{\text{VLB}}(\theta)=\sum_{t=1}^T\mathbb{E}_{x_0, \epsilon}[\mathcal{L}_t(\theta)]$ with
\begin{equation}
    \mathcal{L}_t(\theta)  :=c_t \lVert \epsilon - \epsilon_{\theta_t} (x_t) \rVert^2, \quad c_t := \frac{\beta_t^2}{2\sigma^2_t \alpha_t(1-\bar{\alpha}_t)}\label{eq:Lt}.
\end{equation}
We now have divided the problem into $T$ independent subproblems, where the training with uniform timestep sampling corresponds to uniformly selecting one of these subproblems $\mathcal{L}_{t}$ and performing a stochastic gradient step on its corresponding parameter $\theta_t$. Informally, if certain subproblems are particularly more challenging and require more iterations to optimize, there is a potential for improvement through non-uniform training strategies.

In this sense, we assess the difficulty of these subproblems by measuring a variance of stochastic gradients, $\text{Var}_{x_0, \epsilon}[\nabla_{\theta_t}\mathcal{L}_t(\theta)]$. It is well known that the convergence of stochastic gradient descent is only guaranteed with the learning rates that scale with the variance of the gradient:
\begin{theorem}{\normalfont\textbf{[\citet{garrigos2023handbook}, informal]}}
    \label{theorem:sgd_convergence}
    Given a few reasonable assumptions, for every $\varepsilon > 0$, we can guarantee that $\mathbb{E}\left[ \mathcal{L}_t(\theta_{K}) - \inf \mathcal{L}_t \right] \leq \varepsilon$ provided that
    \begin{align}
        &\gamma = \mathcal{O}\left(\frac{1}{\sqrt{K}}\left(\text{Var}_{x_0, \epsilon}[\nabla_{\theta_t}\mathcal{L}_t(\theta_{*})]\right)^{-\frac{1}{2}}\right), \\
        &K \geq \mathcal{O}\left(\frac{1}{\varepsilon^2}\text{Var}_{x_0, \epsilon}[\nabla_{\theta_t}\mathcal{L}_t(\theta_{*})]\right).
    \end{align}
    Here, $\gamma$ is the learning rate of Stochastic Gradient Descent (SGD), $\theta_{K}$ is the parameter after $K$ steps of gradient updates, and $\theta_{*}$ is the optimal parameter. $K$ is the total number of iterations required to achieve the accuracy $\varepsilon$.
\end{theorem}
In short, \cref{theorem:sgd_convergence} suggests that the learning rate should be scaled inversely proportional to the standard deviation of stochastic gradient. With an appropriately adjusted learning rate, the number of iterations needed for SGD to converge increases in proportion to the gradient's variance. Since modern optimizers like Adam~\citep{kingma2014adam} are designed to ensure convergence by inversely scaling the learning rate with the exponential moving standard deviation, we can infer that the iterations required for a subproblem $\mathcal{L}_t$ to converge will approximately grow in proportion to the variance of the stochastic gradient at the optimal parameter.

\paragraph{Gradient variance of diffusion model} Considering that the gradient variance at each timestep reflects the difficulty of the optimization problem at that specific timestep, we present the measured gradient variance during diffusion model training in the top row of \cref{fig:gradient_variance_kl_divergence}. 
The results clearly indicate an imbalance in gradient variance across timesteps. 
Notably, for all three schedules, the gradient variances at early timesteps are significantly higher than those at later timesteps, and this trend persists without fading quickly.
Since the variance at optimal parameters serves as a lower bound on the number of gradient iterations required for convergence (\cref{theorem:sgd_convergence}), we can anticipate that the early timesteps might pose a bottleneck to training efficiency when timesteps are sampled uniformly. 

The bottom row of \cref{fig:gradient_variance_kl_divergence} shows the actual timestep-wise diffusion loss, and as expected, it shows that the loss near later timesteps decreases substantially faster than the loss near early timesteps. These results suggest that oversampling early timesteps and undersampling later timesteps could enhance overall training efficiency by effectively addressing the bottleneck presented by the subproblems at early timesteps.
\vspace{-0.3cm}
\paragraph{Interdependence of $\mathcal{L}_t$} We have motivated the non-uniform sampling of timesteps by showing that we have varying gradient variance across timesteps, and this leads to a conclusion of non-uniform sampler that samples proportional to the gradient variance. However, without the assumption of independence across subproblems $\mathcal{L}_t$ based on a specific parameterization, the optimality of such a sampler is not guaranteed.

\begin{figure}[t]
    \centering
    \includegraphics[width=1.0
    \linewidth]{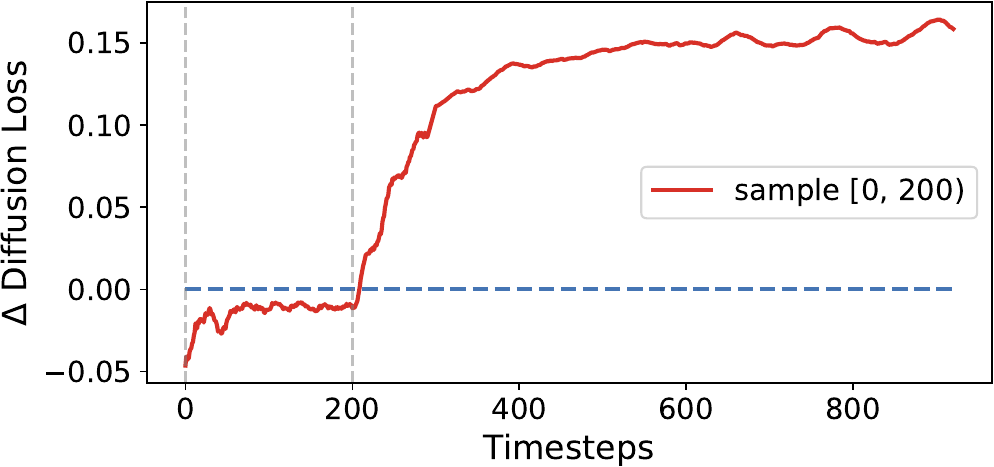}
    \caption{Difference in diffusion loss observed (after and before model update) after training a diffusion model by sampling only timesteps within the range [0, 200). A significant increase in loss is notably observed for the unsampled timesteps.}
    \label{fig:timestep_sampling}
    \vspace{-0.2cm}
\end{figure}

\cref{fig:timestep_sampling} illustrates the extent of interdependence among subproblems during diffusion model training. The figure reveals that the loss increase in untrained timesteps is significantly greater than the loss reduction in the trained timesteps, highlighting the strong interdependence between subproblems. Given that gradient variance can differ by up to a hundredfold (\cref{fig:gradient_variance_kl_divergence}), it is evident that sampling in proportion to the gradient variance alone would likely be insufficient to effectively reduce the loss at later timesteps. Later through experiments, we confirmed that sampling timesteps in proportion to gradient variance yields poor performance~(\cref{tab:variance}).
\section{Adaptive Non-Uniform Timestep Sampling for Diffusion Model Training}
Building on the need for non-uniform timestep sampling scheme  discussed in the previous section, this section introduces an adaptive timestep sampler $\pi$, designed to enhance the training efficiency of diffusion models. We present a method for jointly training this sampler with the diffusion model in an online manner. This approach accelerates the diffusion training process with $\pi$, without requiring prior training data for its optimization.

\subsection{Non-uniform Timestep Sampler}
\label{subsec:non-uniform_timestep_sampler}
We propose a non-uniform timestep sampling strategy for diffusion model training, utilizing a timestep sampler $\pi$ to improve training efficiency:
\begin{align}
\theta &\leftarrow \theta - \gamma \nabla_\theta \mathcal{L}_t,~ t\sim \pi(\cdot|x_0),~\mathcal{L}_t=c_t\lVert \epsilon - \epsilon_\theta (x_t,t) \rVert^2,
\end{align}
where $c_t$ is a schedule-dependent constant as defined in \cref{eq:Lt}. Note that we no longer use independent parameterization across timesteps. 

To leverage the fact that diffusion train loss is similar across closely related timesteps, we parameterize a neural network that generates two positive real scalars, which are then used to sample $t$ from a Beta distribution:
\begin{equation}
    a, b\sim \pi_\phi(\cdot|x_0),\quad a>0,b>0,\quad t\sim \text{Beta}(a, b).
\end{equation}
This ensures our timestep sampler accounts for the high correlation between similar timesteps. For simplicity, we will continue to denote this process as $t\sim \pi_\phi(\cdot|x_0)$.

Our goal is to train the timestep sampler $\pi_\phi$ to minimize the original objective, the variational lower bound $\mathcal{L}_{\text{VLB}}$, as effectively as possible within a finite number of training steps. To achieve this, we seek to optimize the sampler $\pi_{\phi_k}$ to sample timesteps $t$ that yield the greatest reduction of $\mathcal{L}_{\text{VLB}}$ at each iteration $k$. In other words, the sampler $\pi_{\phi_k}$ is trained to maximize the difference in $\mathcal{L}_{\text{VLB}}$ before and after updating the diffusion model at gradient step $k$ using the sampled timestep $t$:
\begin{align}
    \phi_k=&\arg\max_{\phi_k} \mathbb{E}_{x_0, \epsilon, t \sim \pi_{\phi_k}(\cdot|x_0)}\left[\Delta_k^t
    \right], \quad \text{where} \\
     \Delta_k^t=&\frac{1}{T}\sum_{\tau=1}^T \delta_{k, \tau}^t=\frac{1}{T}\sum_{\tau=1}^T\mathcal{L}_{\tau}(\theta_k) - \mathcal{L}_{\tau}\left(\theta_k - \gamma\nabla_{\theta_k}\mathcal{L}_t\right). \nonumber
\end{align}
With such an optimized $\phi_k$, the diffusion model utilizing the sampler $\pi_{\phi_k}$ would be able to adaptively select the best timesteps throughout the training process, minimizing $\mathcal{L}_{\text{VLB}}$ after the update $\theta_k$. In practice, we also include entropy regularization for the sampler $\pi_{\phi}$ in $\Delta_k^t$ to avoid premature convergence.

\paragraph{Look-ahead challenge} 
Evaluating $\mathcal{L}_{\tau}\left(\theta_k - \gamma\nabla_{\theta_k}\mathcal{L}_t\right)$ requires computing how $\theta$ changes after the update, which inherently involves a look-ahead process across different $t$s, making it computationally infeasible. To develop a practical algorithm, we opt for a single gradient update of $\phi_k$, using the last update of $\theta$ as a reference. By applying the likelihood ratio gradient, we use the following update for $\phi_k$:
\begin{equation}
\phi_{k+1}=\phi_k+\gamma \cdot \Delta_k^t \cdot\nabla_{\phi_k}\log \pi_{\phi_k}(\cdot|x_0),
\end{equation}
where we use the identity $\nabla_{\phi} \pi_\phi = \pi_\phi \cdot \nabla_{\phi} \log \pi_\phi$. This is a stochastic approximation using sampled $\epsilon$ and the Beta distribution parameters $a$ and $b$. \cref{alg:training diffusion and policy} outlines the training procedure based on this gradient update for $\phi_k$. Note that $c_t$ is not explicitly included in the algorithm, as it can be effectively subsumed into the sampler $\pi_\phi$ if needed.

\begin{table*}[t!]
    \centering
    \begin{tabular}{cccccc|ccc|cccc}
        \toprule
        & \multicolumn{5}{c}{Linear} & \multicolumn{3}{c}{Cosine} & \multicolumn{4}{c}{Quadratic}\\
        \cmidrule(lr){2-6} \cmidrule(lr){7-9} \cmidrule(lr){10-13}
        & 0.2M & 0.4M & 0.6M & 0.8M & 1.0M & 0.2M & 0.4M & 0.6M & 0.2M & 0.4M & 0.6M &0.8M\\
        \midrule
        DDPM~\citep{ho2020denoising}   &8.16  &6.03  &5.37  &4.28  &3.19 &3.74 &3.14 & 3.14 &4.04 &4.32 &3.01 &3.01\\
        P2~\citep{choi2022perception}     & 6.45 & 5.21 & 4.83 & 4.70 & 4.70 &5.64  &5.21  &5.04  & 4.93 &4.06 &3.74 &3.66\\
        Min-SNR~\citep{hang2023efficient}& 4.40 & 3.51 & 3.28 & 3.16 & 3.16  &\textbf{3.51} &3.07 & 3.07 &4.76 &4.17 &4.17 &4.17\\
        Log-normal~\citep{karras2022elucidating} &5.38 &4.36 &3.92 &3.60 &3.60 &4.55 &3.89 &3.78 &4.78 &3.81 &3.61 &3.51 \\
        SpeeD~\citep{wang2024closer} &4.08 &3.66 &3.57 &3.57 &3.57 &4.12 &3.57 &3.57 &3.99 &3.54 &3.52 &3.52 \\
        \midrule
        Ours   & \textbf{3.99} & \textbf{3.21} & \textbf{3.10} & \textbf{2.94} & \textbf{2.94} & 3.87 & \textbf{3.06} & \textbf{3.00} & \textbf{3.96} & \textbf{3.07} & \textbf{2.95} & \textbf{2.95}  \\
        \bottomrule
    \end{tabular}
    \caption{Performance (FID) comparison on the CIFAR-10 dataset across different acceleration methods utilizing Linear, Cosine, and Quadratic schedules. We reported the lowest FID score achieved for each training duration. The algorithm achieving the lowest FID score is highlighted in bold.
    }
    \label{tab:combined_table}
\end{table*}

\begin{table}[t!]
    \centering
    \begin{tabular}{ccc}\toprule
         &100 epochs  
         &300 epochs\\
         \midrule
         LDM & 12.87 
         & 8.72 \\
         P2 & \textbf{10.69} 
         & 6.88 \\
         Min-SNR & 10.87 
         & 6.83 \\
         Log-normal & 12.81 
         & 7.28 \\
         SpeeD & 14.40 
         & 7.98 \\
         \midrule
         Ours & 10.94 
         & \textbf{6.62} \\
         \bottomrule
    \end{tabular}
    \caption{Performance (FID) comparison on the CelebA-HQ 256 $\times$ 256 dataset across different acceleration methods.~The algorithm achieving the lowest FID score is highlighted in bold.
    \vspace{-0.3cm}}
    \label{tab:celeba-hq}
\end{table}

\begin{algorithm}[t!]
\caption{Training DM with Timestep Sampler}
\label{alg:training diffusion and policy}
\textbf{Parameter}: $\epsilon_\theta$, $\pi_\phi$

\begin{algorithmic}[1] 
\FOR{gradient step $k$}
\STATE $x_0 \sim q(x_0)$, $\epsilon \sim \mathcal{N}(0, I)$, \STATE $a, b \sim \pi_{\phi_k}(\cdot|x_0)$, $t \sim \text{Beta}(a, b)$
\STATE $x_t = \sqrt{\bar{\alpha}_t}x_0 + \sqrt{1- \bar{\alpha}_t} \epsilon$
\STATE $\theta_{k+1} = \theta_k - \gamma \cdot \nabla_{\theta_k} \lVert \epsilon - \epsilon_{\theta_{k}} (x_t,t) \rVert^2$
\IF{$k \bmod f_S = 0$}
\STATE $\Tilde{\Delta}_k^t =$ approximate $\Delta_k^t$ with \cref{alg:reward approximation}
\STATE $\phi_{k+1} = \phi_k + \gamma \cdot \Tilde{\Delta}_k^t \nabla_{\phi_k}\log \pi_{\phi_k}(a, b|x_0)$
\ENDIF
\ENDFOR
\end{algorithmic}
\end{algorithm}

\begin{algorithm}[t]
\caption{Approximation of $\Delta_k^t$}
\label{alg:reward approximation}
\textbf{Input}: $\theta_k$, $\theta_{k+1}$

\textbf{Initialize}: fixed-size queue $Q$

\begin{algorithmic}[1] 
\STATE Sample one $x_0$ randomly from dataset
\STATE Calculate $\{\delta_{k, \tau}^t\}_{\tau=1}^T$  for the sampled $x_0$
\STATE Push $\{\delta_{k, \tau}^t\}_{\tau=1}^T$ into queue $Q$
\IF{$|Q| > 1$}
\STATE Determine the set of timesteps $S$ using a feature selection method with queue $Q$
\ENDIF
\RETURN $\Tilde{\Delta}^t_k = \frac{1}{|S|}\sum_{\tau \in S} \delta_{k, \tau}^t$ for $x_0$s in current mini-batch 
\end{algorithmic}
\end{algorithm}

\subsection{Approximation of $\Delta_k^t$}
Even with the reduced computation offered by the surrogate algorithm described above, evaluating $\Delta_k^t$ still requires calculating the exact objective $\mathcal{L}_{\text{VLB}}$ before and after updating $\theta_k$, i.e., $\Delta_k^t=\mathcal{L}_{\text{VLB}}(\theta_k)-\mathcal{L}_{\text{VLB}}(\theta_{k+1})$. Fortunately, by leveraging the correlation between the losses of nearby timesteps, we can accurately approximate $\Delta_k^t$ using a small set of timesteps $S\subset \{t\}_{t=1}^T$, i.e., 
\begin{equation}
    {\Delta}_k^t \approx \tilde{\Delta}_k^t =\frac{1}{|S|}\sum_{\tau\in S} \delta_{k, \tau}^t= \frac{1}{|S|}\sum_{\tau \in S} \mathcal{L}_{\tau}(\theta_k)-\mathcal{L}_{\tau}(\theta_{k+1}).
\end{equation}
To determine $S$, for each $k$, we calculate $\{\delta_{k, \tau}^t\}_{\tau=1}^T$ for a single $x_0$ and store these values in a queue.
Using any off-the-shelf feature selection algorithm, we then identify the top $|S|$ timesteps that best explain $\Delta_k^t$ based on the data from the queue, with $|S|$ being a hyperparameter. We mainly used a very small subset $|S|=3$ to approximate $\Delta_k^t$. Later through experiments, we demonstrate that this small subset is sufficient for achieving strong performance with respect to wall-clock time (\cref{fig:compare_ablation}). 

The process of approximating $\Delta_k^t$ using this feature selection approach is described in \cref{alg:reward approximation}. Note in practice, we further reduce the computational burden by updating $\phi$ only once per $f_S$ gradient iterations of the diffusion model. The overall structure of our algorithms is in \cref{fig:overall_algorithm}. The computational overhead introduced by the proposed timestep sampling mechanism is in Appendix \cref{sec:Timecomplexity}.
\begin{figure*}[t!]
    \centering
    \includegraphics[width=1.0\linewidth]{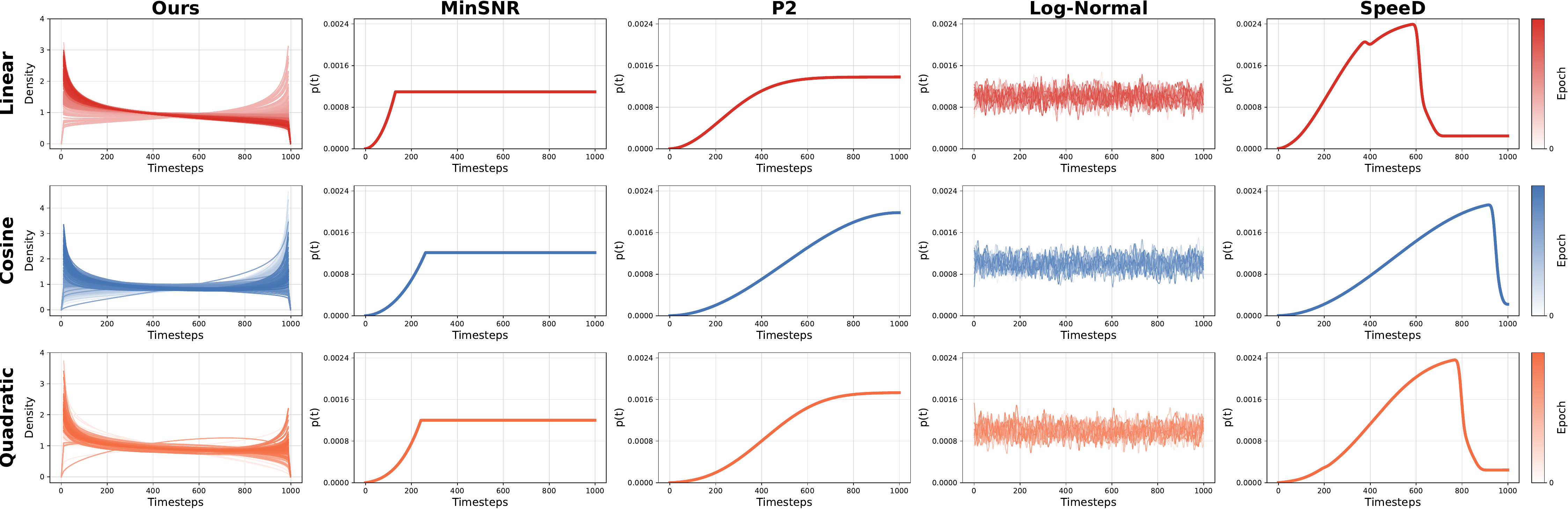}
    \caption{Visualization of timestep sampling schemes of our method, Min-SNR, P2, Log-Normal, and SpeeD. For weighting methods (Min-SNR, P2), weights at each timestep are converted to probabilities. The color bar on the right represents the progression of epochs, providing a view of how each scheduling method samples timesteps across the training process.}
    \label{fig:viz_timestep}
\end{figure*}

\section{Experiments} 
In all experiments, we trained each diffusion architecture with its official implementation. Simultaneously, the timestep sampler $\pi_{\phi}$ was trained using $\Delta_k^t$, which approximates the change in $\mathcal{L}_{\text{VLB}}$ before and after updating the diffusion model. We begin by evaluating the effectiveness of our learning method in comparison to heuristic approaches in \cref{subsec:comparison_to_heuristics}. Then we assess the compatibility of our model with heuristic approaches in \cref{subsec:compatibility_with_heuristics}. Finally, we present additional analyses to further support our method in \cref{subsec:analysis}.

\paragraph{Experimental Setup} Frechet Inception Distance (FID) \citep{heusel2017gans} is used to evaluate both the fidelity and coverage of generated images. We mainly used a machine equipped with NVIDIA RTX 4090 GPUs to train the models. Detailed hyperparameter values and implementation details can be found in Appendix \cref{sec:implementation_details} and \cref{sec:hyperparameters}.

\subsection{Comparison to Heuristic Acceleration Methods}
\label{subsec:comparison_to_heuristics}

\paragraph{Comparison across various noise schedules} 
To compare our model and other acceleration techniques across various noise schedules, we used the CIFAR-10 32$\times$32 dataset~\citep{alex2009learning}, utilizing DDPM~\citep{ho2020denoising} as the backbone architecture. 

\cref{tab:combined_table} demonstrates that our method, despite its learning-based nature, generally achieves faster convergence and better performance at each iteration compared to heuristic approaches \citep{choi2022perception, hang2023efficient, karras2022elucidating, wang2024closer}. Moreover, our method demonstrates consistent robustness across different noise schedules. In contrast, while Min-SNR performs the best among heuristic approaches with linear and cosine noise schedules, its performance declines significantly with a quadratic noise schedule. As illustrated in \cref{fig:viz_timestep}, our method achieves robust performance by adaptively sampling timesteps throughout the training process, whereas heuristic approaches rely on fixed sampling strategies. Furthermore, \cref{fig:main_wallclocktime} shows that our method also achieves faster convergence in terms of wall-clock time compared to heuristic approaches.

\paragraph{Comparison on high resolution images} 
To evaluate the acceleration methods on high-resolution images, we used the CelebA-HQ 256$\times$256 dataset \citep{karras2018progressive}, utilizing LDM~\citep{rombach2022high} as the backbone architecture. \cref{tab:celeba-hq} demonstrates that although heuristic methods initially achieve better performance, our learning-based approach surpasses them over time. We also found that P2 performs comparably well in this setting relative to the results in \cref{tab:combined_table}, suggesting that the performance of heuristics can vary across different settings. Visualizations of the generated CelebA-HQ images from our method and baselines are in Appendix \cref{sec:Visualization}.

\subsection{Compatibility with Heuristic Approaches}
\label{subsec:compatibility_with_heuristics}
In \cref{subsec:comparison_to_heuristics}, we demonstrated that while our learning-based method initially accelerates more slowly than the heuristic approaches due to its learning-based nature, it ultimately converges to a point with better optimality in later iterations. However, we found that heuristic weighting schemes can be incorporated into our model to achieve faster convergence from the beginning as well. Indeed, heuristic weightings can be regarded as part of the schedule-dependent constant $c_t$, and can be effectively subsumed into our sampler $\pi_\phi$ (\cref{subsec:non-uniform_timestep_sampler}). Thus, our sampler can effectively account for the additional weight incorporated into the diffusion loss through heuristic approaches.

To assess the compatibility of our model with heuristic approaches, we trained our model along with Min-SNR weighting on the ImageNet 256$\times$256 benchmark \citep{deng2009imagenet}. 
We used the encoder from LDM~\citep{rombach2022high} to convert images of dimensions $ 256 \times 256 \times 3 $ into latent representations of size $ 32 \times 32 \times 4 $. 
We used U-ViT \citep{bao2023all} as our diffusion backbone. \cref{tab:imagnet} shows that our method with heuristic weighting can achieve faster convergence from the beginning as well, ultimately converging to a lower FID score compared to the baseline. 
\begin{figure*}[t]
    \centering
    \includegraphics[width=0.66\linewidth]{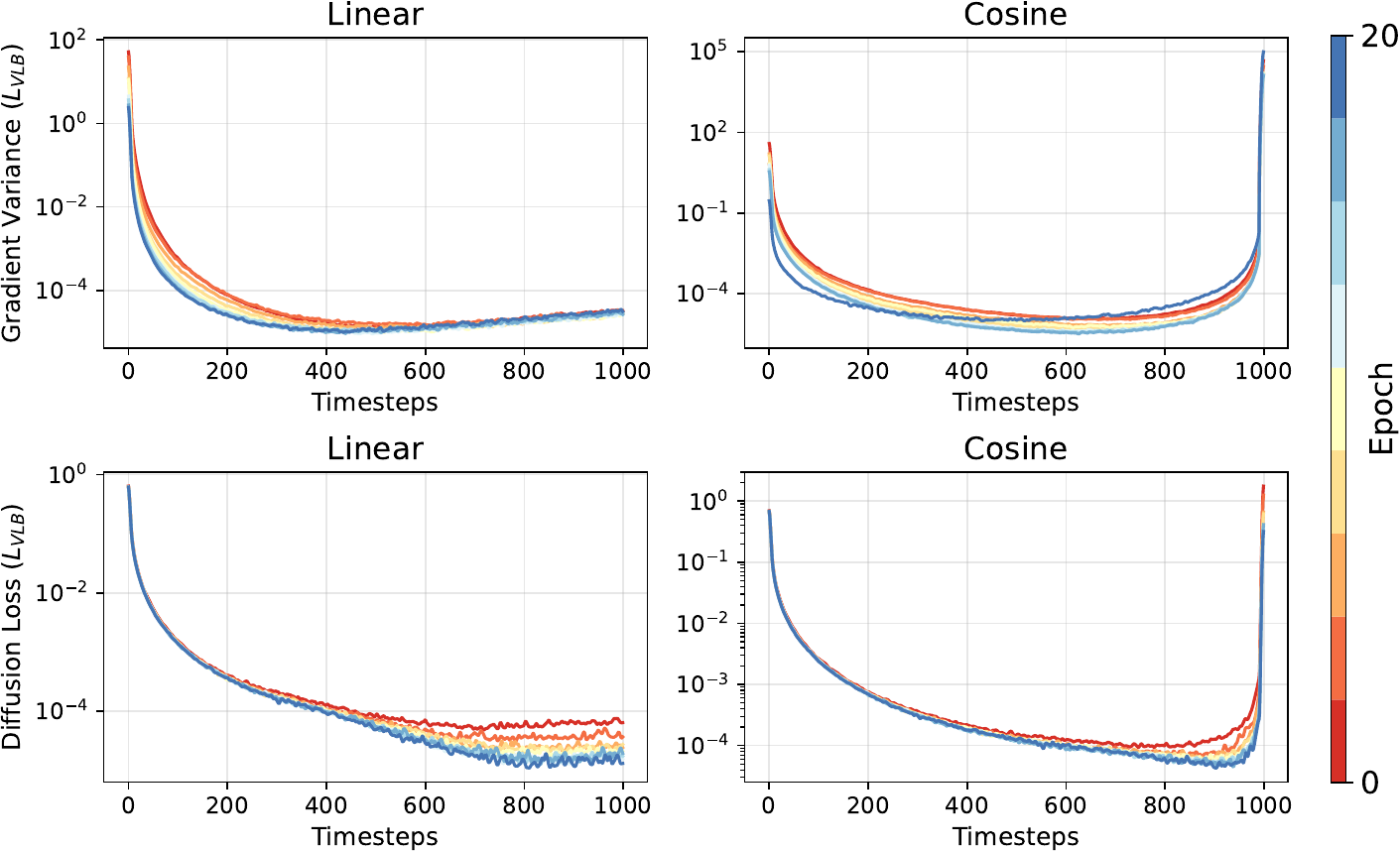}
    \includegraphics[width=0.314\linewidth]{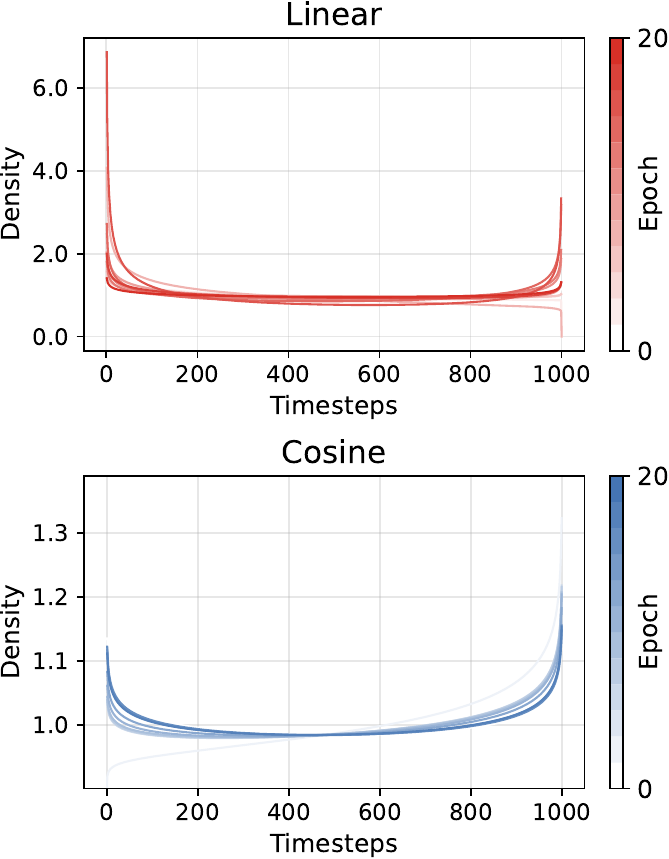}    
    \caption{\textbf{(Left, Center)} Comparison of gradient variance (top) and diffusion loss (bottom) over timesteps for linear and cosine schedules. The color bar on the right indicates the progression of epochs. Unlike the results presented in \cref{fig:gradient_variance_kl_divergence}, these results are based on training DDPM~\citep{ho2020denoising} with $\mathcal{L}_{\text{VLB}}$ for 20 epochs. \textbf{(Right)} Visualization of our model's timestep distributions over training, with a linear schedule (top) and a cosine schedule (bottom).}
    \label{fig:adaptive_sampling}
\end{figure*}

\begin{table}[t!]
    \centering
    \begin{tabular}{ccccccc}
         \toprule
         &0.2M &0.4M &0.6M &0.8M &1.0M \\
         \midrule
         Min-SNR &11.5 &6.73 &5.28 &4.80 &4.54 \\
         \midrule
         Min-SNR  &\multirow{2}{*}{\textbf{11.3}} &\multirow{2}{*}{\textbf{6.59}} &\multirow{2}{*}{\textbf{5.19}} &\multirow{2}{*}{\textbf{4.66}} &\multirow{2}{*}{\textbf{4.33}}\\
         +Ours \\
         \bottomrule
    \end{tabular}
    \caption{Performance (FID) comparison on the ImageNet 256 $\times$ 256 dataset between Min-SNR and our method with Min-SNR. The algorithm achieving the lowest FID score is highlighted in bold.}
    \label{tab:imagnet}
\end{table}

\begin{table}[t!]
    \centering
    \begin{tabular}{cccccc}\toprule
         &0.2M &0.4M &0.6M &0.8M &1.0M \\
         \midrule
         Variance &21.3 &16.0 &13.5 &12.1 &11.3\\
         \midrule
         Ours &\textbf{3.99} & \textbf{3.21} & \textbf{3.10} & \textbf{2.94} & \textbf{2.94} \\
         \bottomrule
    \end{tabular}
    \caption{Performance (FID) comparison on the CIFAR-10 dataset between our method and the approach that samples timesteps proportional to gradient variance. We reported the lowest FID score achieved for each training duration. The algorithm achieving the lowest FID score is highlighted in bold. 
    }
    \vspace{-0.3cm}
    \label{tab:variance}
\end{table}

\begin{figure}[t]
    \centering
    \includegraphics[width=1.0\linewidth]{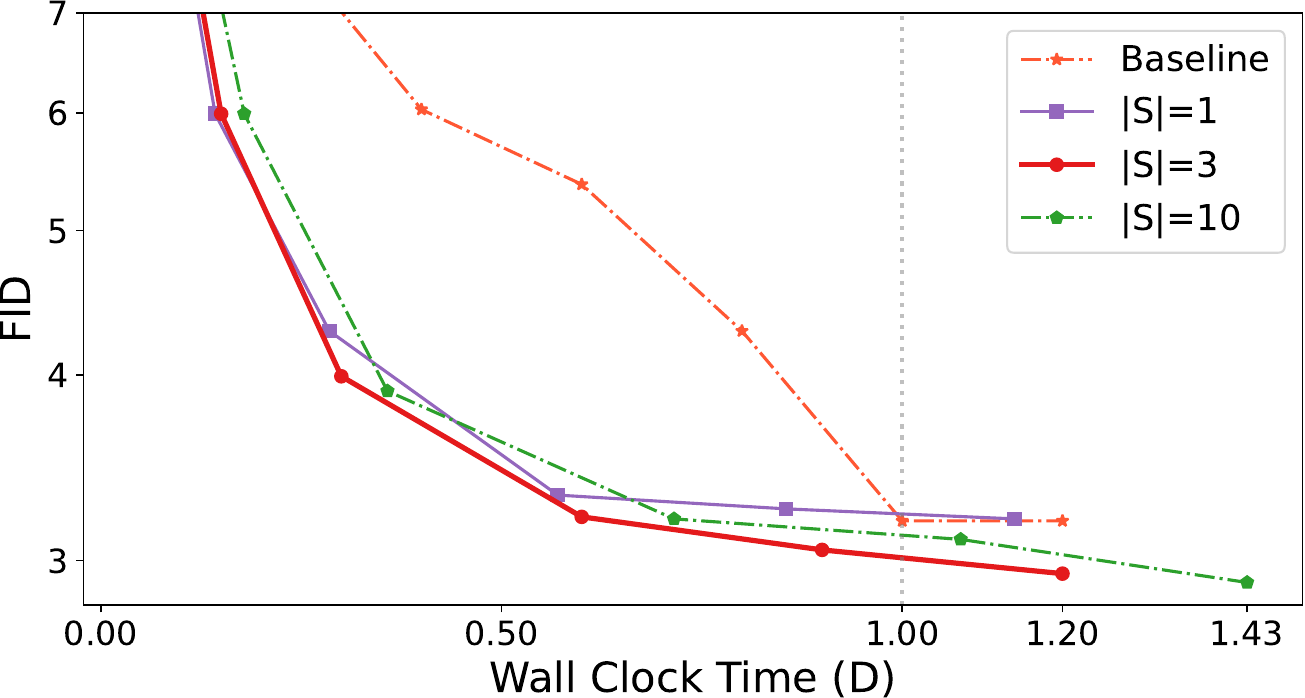}
    \caption{Comparison of FID scores of baseline and our method against relative wall clock time, where 1D represents the time it takes for the baseline to converge.~$|S|$ denotes the number of timesteps used in our model to approximate $\Delta_k^t$.}
    \label{fig:compare_ablation}
    \vspace{-0.3cm}
\end{figure}

\subsection{Analysis}
\label{subsec:analysis}
\paragraph{Analysis of non-uniform timestep sampling} 

\cref{fig:adaptive_sampling} (right column) illustrates the timestep distribution sampled by our method when applied to DDPM. \cref{fig:adaptive_sampling} (left, center column) illustrates the variations in gradient variance and diffusion loss observed when training DDPM using $\mathcal{L}_{\text{VLB}}$ rather than $\mathcal{L}_{\text{DDPM}}$, without incorporating our proposed method. Training with $\mathcal{L}_{\text{VLB}}$ differs from the conventional diffusion training with $\mathcal{L}_{\text{DDPM}}$ (\cref{fig:gradient_variance_kl_divergence}) in that $\mathcal{L}_{\text{VLB}}$ weights the losses across timesteps differently using $c_t$. Since $c_t$ has larger values at later timesteps, it leads to increased loss and gradient variance at these timesteps, especially with the cosine schedule as shown in \cref{fig:adaptive_sampling}. 

In \cref{section: Diffusion Model Training with Adam Optimizer}, we argued that the number of iterations required for convergence would be proportional to the gradient variance at convergence, assuming independence across timesteps. The timestep sampling distributions shown in the plots support this intuition, with the linear schedule heavily sampling low timesteps and the cosine schedule sampling predominantly from both low and high timesteps. This suggests that the direction that most effectively reduces the original objective, $\mathcal{L}_{\text{VLB}}$, somewhat aligns with sampling timesteps that exhibit higher gradient variance. On the other hand, \cref{tab:variance} demonstrates that sampling timesteps in proportion to gradient variance results in poor performance, likely due to the interdependencies among the subproblems $\mathcal{L}_t$. 
\vspace{-0.2cm}

\paragraph{Ablation study on $|S|$}
\cref{fig:compare_ablation} illustrates the effect of varying the number of selected timesteps $|S|$ on the performance of our proposed model. All choices of $|S|$ show acceleration over the baseline, suggesting that we can approximate $\Delta_k^t$ using only a small subset of timesteps. We observed that $|S|=3$ achieves the best performance in terms of wall clock time. Thus, we primarily used $|S|=3$ to approximate $\Delta_k^t$ throughout our experiments.

\section{Conclusion}
In this paper, we discovered the variations in stochastic gradient variance across different diffusion training timesteps and argued that such variation can potentially lead to slow convergence. However, the strong interdependence of gradients across different timesteps suggests that this insight, on its own, does not directly result in optimal acceleration of diffusion training. To this end, we propose a more direct approach that estimates the impact of gradient updates on the variational lower bound for each timestep and adjusts the sampling frequency to prioritize timesteps that require further optimization. Experimental results demonstrate that our method significantly accelerates the convergence of the diffusion training process compared to previous heuristics, while also exhibiting robust performance across various datasets, scheduling strategies, and diffusion architectures. 

\paragraph{Limitations and future work}
Our learning-based method incurs higher computational costs compared to heuristic approaches, with resource demands increasing significantly in larger problem domains. Furthermore, we have not yet explored the application of our method within the score-based diffusion framework, which combines score-based generative modeling with diffusion probabilistic modeling and shows notable results in image generation tasks \citep{song2020score, wu2023fast, karras2022elucidating}. Addressing these limitations in terms of computational efficiency and extending the method's applicability within score-based diffusion models remains a promising direction for future research.

\section{Acknowledgements}
This work was partly supported by Institute of Information $\&$ communications Technology Planning $\&$ Evaluation (IITP) grant funded by the Korea government (MSIT) (No. RS-2022-II220311, Development of Goal-Oriented Reinforcement Learning Techniques for Contact-Rich Robotic Manipulation of Everyday Objects, No. RS-2024-00457882, AI Research Hub Project, and No. RS-2019-II190079, Artificial Intelligence Graduate School Program (Korea University)), the IITP(Institute of Information $\&$ Coummunications Technology Planning $\&$ Evaluation)-ITRC(Information Technology Research Center)(IITP-2024-RS-2024-00436857) grant funded by the Korea government (Ministry of Science and ICT), the NRF (RS-2024-00451162) funded by the Ministry of Science and ICT, Korea, BK21 Four project of the National Research Foundation of Korea, and the National Research Foundation of Korea (NRF) grant funded by the Korea government (MSIT)(RS-2025-00560367).


{
    \small
    \bibliographystyle{ieeenat_fullname}
    \bibliography{main}

\begin{thebibliography}{35}
\providecommand{\natexlab}[1]{#1}
\providecommand{\url}[1]{\texttt{#1}}
\expandafter\ifx\csname urlstyle\endcsname\relax
  \providecommand{\doi}[1]{doi: #1}\else
  \providecommand{\doi}{doi: \begingroup \urlstyle{rm}\Url}\fi

\bibitem[Ajay et~al.(2022)Ajay, Du, Gupta, Tenenbaum, Jaakkola, and Agrawal]{ajay2022conditional}
Anurag Ajay, Yilun Du, Abhi Gupta, Joshua Tenenbaum, Tommi Jaakkola, and Pulkit Agrawal.
\newblock Is conditional generative modeling all you need for decision-making?
\newblock \emph{arXiv preprint arXiv:2211.15657}, 2022.

\bibitem[Alex(2009)]{alex2009learning}
Krizhevsky Alex.
\newblock Learning multiple layers of features from tiny images.
\newblock \emph{https://www. cs. toronto. edu/kriz/learning-features-2009-TR. pdf}, 2009.

\bibitem[Andrychowicz et~al.(2016)Andrychowicz, Denil, Gomez, Hoffman, Pfau, Schaul, Shillingford, and De~Freitas]{andrychowicz2016learning}
Marcin Andrychowicz, Misha Denil, Sergio Gomez, Matthew~W Hoffman, David Pfau, Tom Schaul, Brendan Shillingford, and Nando De~Freitas.
\newblock Learning to learn by gradient descent by gradient descent.
\newblock \emph{Advances in neural information processing systems}, 29, 2016.

\bibitem[Austin et~al.(2021)Austin, Johnson, Ho, Tarlow, and Van Den~Berg]{austin2021structured}
Jacob Austin, Daniel~D Johnson, Jonathan Ho, Daniel Tarlow, and Rianne Van Den~Berg.
\newblock Structured denoising diffusion models in discrete state-spaces.
\newblock \emph{Advances in Neural Information Processing Systems}, 34:\penalty0 17981--17993, 2021.

\bibitem[Bao et~al.(2023)Bao, Nie, Xue, Cao, Li, Su, and Zhu]{bao2023all}
Fan Bao, Shen Nie, Kaiwen Xue, Yue Cao, Chongxuan Li, Hang Su, and Jun Zhu.
\newblock All are worth words: A vit backbone for diffusion models.
\newblock In \emph{Proceedings of the IEEE/CVF conference on computer vision and pattern recognition}, pages 22669--22679, 2023.

\bibitem[Bello et~al.(2017)Bello, Zoph, Vasudevan, and Le]{bello2017neural}
Irwan Bello, Barret Zoph, Vijay Vasudevan, and Quoc~V Le.
\newblock Neural optimizer search with reinforcement learning.
\newblock In \emph{International Conference on Machine Learning}, pages 459--468. PMLR, 2017.

\bibitem[Choi et~al.(2022)Choi, Lee, Shin, Kim, Kim, and Yoon]{choi2022perception}
Jooyoung Choi, Jungbeom Lee, Chaehun Shin, Sungwon Kim, Hyunwoo Kim, and Sungroh Yoon.
\newblock Perception prioritized training of diffusion models.
\newblock In \emph{Proceedings of the IEEE/CVF Conference on Computer Vision and Pattern Recognition}, pages 11472--11481, 2022.

\bibitem[Deng et~al.(2009)Deng, Dong, Socher, Li, Li, and Fei-Fei]{deng2009imagenet}
Jia Deng, Wei Dong, Richard Socher, Li-Jia Li, Kai Li, and Li Fei-Fei.
\newblock Imagenet: A large-scale hierarchical image database.
\newblock In \emph{2009 IEEE conference on computer vision and pattern recognition}, pages 248--255. Ieee, 2009.

\bibitem[Dhariwal and Nichol(2021)]{dhariwal2021diffusion}
Prafulla Dhariwal and Alexander Nichol.
\newblock Diffusion models beat gans on image synthesis.
\newblock \emph{Advances in neural information processing systems}, 34:\penalty0 8780--8794, 2021.

\bibitem[Garrigos and Gower(2023)]{garrigos2023handbook}
Guillaume Garrigos and Robert~M Gower.
\newblock Handbook of convergence theorems for (stochastic) gradient methods.
\newblock \emph{arXiv preprint arXiv:2301.11235}, 2023.

\bibitem[Hang et~al.(2023)Hang, Gu, Li, Bao, Chen, Hu, Geng, and Guo]{hang2023efficient}
Tiankai Hang, Shuyang Gu, Chen Li, Jianmin Bao, Dong Chen, Han Hu, Xin Geng, and Baining Guo.
\newblock Efficient diffusion training via min-snr weighting strategy.
\newblock In \emph{Proceedings of the IEEE/CVF International Conference on Computer Vision}, pages 7441--7451, 2023.

\bibitem[He et~al.(2022)He, Sun, Wang, Huang, and Qiu]{he2022diffusionbert}
Zhengfu He, Tianxiang Sun, Kuanning Wang, Xuanjing Huang, and Xipeng Qiu.
\newblock Diffusionbert: Improving generative masked language models with diffusion models.
\newblock \emph{arXiv preprint arXiv:2211.15029}, 2022.

\bibitem[Heusel et~al.(2017)Heusel, Ramsauer, Unterthiner, Nessler, and Hochreiter]{heusel2017gans}
Martin Heusel, Hubert Ramsauer, Thomas Unterthiner, Bernhard Nessler, and Sepp Hochreiter.
\newblock Gans trained by a two time-scale update rule converge to a local nash equilibrium.
\newblock \emph{Advances in neural information processing systems}, 30, 2017.

\bibitem[Ho et~al.(2020)Ho, Jain, and Abbeel]{ho2020denoising}
Jonathan Ho, Ajay Jain, and Pieter Abbeel.
\newblock Denoising diffusion probabilistic models.
\newblock \emph{Advances in neural information processing systems}, 33:\penalty0 6840--6851, 2020.

\bibitem[Janner et~al.(2022)Janner, Du, Tenenbaum, and Levine]{janner2022planning}
Michael Janner, Yilun Du, Joshua~B Tenenbaum, and Sergey Levine.
\newblock Planning with diffusion for flexible behavior synthesis.
\newblock \emph{arXiv preprint arXiv:2205.09991}, 2022.

\bibitem[Karras et~al.(2018)Karras, Aila, Laine, and Lehtinen]{karras2018progressive}
Tero Karras, Timo Aila, Samuli Laine, and Jaakko Lehtinen.
\newblock Progressive growing of gans for improved quality, stability, and variation. arxiv 2017.
\newblock \emph{arXiv preprint arXiv:1710.10196}, pages 1--26, 2018.

\bibitem[Karras et~al.(2022)Karras, Aittala, Aila, and Laine]{karras2022elucidating}
Tero Karras, Miika Aittala, Timo Aila, and Samuli Laine.
\newblock Elucidating the design space of diffusion-based generative models.
\newblock \emph{Advances in neural information processing systems}, 35:\penalty0 26565--26577, 2022.

\bibitem[Khachatryan et~al.(2023)Khachatryan, Movsisyan, Tadevosyan, Henschel, Wang, Navasardyan, and Shi]{khachatryan2023text2video}
Levon Khachatryan, Andranik Movsisyan, Vahram Tadevosyan, Roberto Henschel, Zhangyang Wang, Shant Navasardyan, and Humphrey Shi.
\newblock Text2video-zero: Text-to-image diffusion models are zero-shot video generators.
\newblock In \emph{Proceedings of the IEEE/CVF International Conference on Computer Vision}, pages 15954--15964, 2023.

\bibitem[Kingma and Ba(2014)]{kingma2014adam}
Diederik~P Kingma and Jimmy Ba.
\newblock Adam: A method for stochastic optimization.
\newblock \emph{arXiv preprint arXiv:1412.6980}, 2014.

\bibitem[Li and Malik(2016)]{li2016learning}
Ke Li and Jitendra Malik.
\newblock Learning to optimize.
\newblock \emph{arXiv preprint arXiv:1606.01885}, 2016.

\bibitem[Li et~al.(2020)Li, Zhao, Varma, Salpekar, Noordhuis, Li, Paszke, Smith, Vaughan, Damania, et~al.]{li2020pytorch}
Shen Li, Yanli Zhao, Rohan Varma, Omkar Salpekar, Pieter Noordhuis, Teng Li, Adam Paszke, Jeff Smith, Brian Vaughan, Pritam Damania, et~al.
\newblock Pytorch distributed: Experiences on accelerating data parallel training.
\newblock \emph{arXiv preprint arXiv:2006.15704}, 2020.

\bibitem[Li et~al.(2022)Li, Thickstun, Gulrajani, Liang, and Hashimoto]{li2022diffusion}
Xiang Li, John Thickstun, Ishaan Gulrajani, Percy~S Liang, and Tatsunori~B Hashimoto.
\newblock Diffusion-lm improves controllable text generation.
\newblock \emph{Advances in Neural Information Processing Systems}, 35:\penalty0 4328--4343, 2022.

\bibitem[Nichol et~al.(2021)Nichol, Dhariwal, Ramesh, Shyam, Mishkin, McGrew, Sutskever, and Chen]{nichol2021glide}
Alex Nichol, Prafulla Dhariwal, Aditya Ramesh, Pranav Shyam, Pamela Mishkin, Bob McGrew, Ilya Sutskever, and Mark Chen.
\newblock Glide: Towards photorealistic image generation and editing with text-guided diffusion models.
\newblock \emph{arXiv preprint arXiv:2112.10741}, 2021.

\bibitem[Rombach et~al.(2022)Rombach, Blattmann, Lorenz, Esser, and Ommer]{rombach2022high}
Robin Rombach, Andreas Blattmann, Dominik Lorenz, Patrick Esser, and Bj{\"o}rn Ommer.
\newblock High-resolution image synthesis with latent diffusion models.
\newblock In \emph{Proceedings of the IEEE/CVF conference on computer vision and pattern recognition}, pages 10684--10695, 2022.

\bibitem[Sanokowski et~al.(2024)Sanokowski, Hochreiter, and Lehner]{sanokowski2024diffusion}
Sebastian Sanokowski, Sepp Hochreiter, and Sebastian Lehner.
\newblock A diffusion model framework for unsupervised neural combinatorial optimization.
\newblock \emph{arXiv preprint arXiv:2406.01661}, 2024.

\bibitem[Schneider(2023)]{schneider2023archisound}
Flavio Schneider.
\newblock Archisound: Audio generation with diffusion.
\newblock \emph{arXiv preprint arXiv:2301.13267}, 2023.

\bibitem[Sohl-Dickstein et~al.(2015)Sohl-Dickstein, Weiss, Maheswaranathan, and Ganguli]{sohl2015deep}
Jascha Sohl-Dickstein, Eric Weiss, Niru Maheswaranathan, and Surya Ganguli.
\newblock Deep unsupervised learning using nonequilibrium thermodynamics.
\newblock In \emph{International conference on machine learning}, pages 2256--2265. PMLR, 2015.

\bibitem[Song et~al.(2020{\natexlab{a}})Song, Meng, and Ermon]{song2020denoising}
Jiaming Song, Chenlin Meng, and Stefano Ermon.
\newblock Denoising diffusion implicit models.
\newblock \emph{arXiv preprint arXiv:2010.02502}, 2020{\natexlab{a}}.

\bibitem[Song et~al.(2020{\natexlab{b}})Song, Sohl-Dickstein, Kingma, Kumar, Ermon, and Poole]{song2020score}
Yang Song, Jascha Sohl-Dickstein, Diederik~P Kingma, Abhishek Kumar, Stefano Ermon, and Ben Poole.
\newblock Score-based generative modeling through stochastic differential equations.
\newblock \emph{arXiv preprint arXiv:2011.13456}, 2020{\natexlab{b}}.

\bibitem[Sun and Yang(2023)]{sun2023difusco}
Zhiqing Sun and Yiming Yang.
\newblock Difusco: Graph-based diffusion solvers for combinatorial optimization.
\newblock \emph{Advances in Neural Information Processing Systems}, 36:\penalty0 3706--3731, 2023.

\bibitem[Wang et~al.(2024)Wang, Zhou, Shi, Yuan, Shang, Peng, Zhang, and You]{wang2024closer}
Kai Wang, Yukun Zhou, Mingjia Shi, Zhihang Yuan, Yuzhang Shang, Xiaojiang Peng, Hanwang Zhang, and Yang You.
\newblock A closer look at time steps is worthy of triple speed-up for diffusion model training.
\newblock \emph{arXiv preprint arXiv:2405.17403}, 2024.

\bibitem[Wang et~al.(2022)Wang, Hunt, and Zhou]{wang2022diffusion}
Zhendong Wang, Jonathan~J Hunt, and Mingyuan Zhou.
\newblock Diffusion policies as an expressive policy class for offline reinforcement learning.
\newblock \emph{arXiv preprint arXiv:2208.06193}, 2022.

\bibitem[Wichrowska et~al.(2017)Wichrowska, Maheswaranathan, Hoffman, Colmenarejo, Denil, Freitas, and Sohl-Dickstein]{wichrowska2017learned}
Olga Wichrowska, Niru Maheswaranathan, Matthew~W Hoffman, Sergio~Gomez Colmenarejo, Misha Denil, Nando Freitas, and Jascha Sohl-Dickstein.
\newblock Learned optimizers that scale and generalize.
\newblock In \emph{International conference on machine learning}, pages 3751--3760. PMLR, 2017.

\bibitem[Wu et~al.(2023)Wu, Zhou, Kawaguchi, and Zhang]{wu2023fast}
Zike Wu, Pan Zhou, Kenji Kawaguchi, and Hanwang Zhang.
\newblock Fast diffusion model.
\newblock \emph{arXiv preprint arXiv:2306.06991}, 2023.

\bibitem[Zheng et~al.(2024)Zheng, Peng, and You]{zheng2024open}
Zangwei Zheng, Xiangyu Peng, and Yang You.
\newblock Open-sora: Democratizing efficient video production for all, 2024.

\end{thebibliography}
}
\clearpage
\setcounter{page}{1}
\onecolumn
{\centering
\Large
\textbf{\thetitle}\\
\vspace{0.5em}Supplementary Material \\
\vspace{1.0em}}

\section{Time Complexity of Our Algorithm}
\label{sec:Timecomplexity}
We compute the additional computational overhead introduced by the proposed timestep sampling mechanism. During each iteration of diffusion model training, particularly for sufficiently large models, the majority of the time is dedicated to the forward and backward passes of the objective $\mathcal{L}_t(\theta)$. The backward pass, in particular, is more computationally demanding due to the communication step required for gradient synchronization, typically taking about three times longer than the forward pass~\citep{li2020pytorch}. Thus, if we denote the time required for the forward pass on a batch of data $B$ as $t_{\text{fwd}(\theta)}$, a single training iteration will approximately take:
\begin{equation*}
t_{\text{iter}(\theta)}=t_{\text{fwd}(\theta)} + t_{\text{bwd}(\theta)}\approx 4\cdot t_{\text{fwd}(\theta)}.
\end{equation*}
If we do not approximate $\Delta_k^t$, its computation requires evaluating $\sum_{t=1}^T\mathcal{L}_t(\cdot)$ twice---once before and once after updating $\theta$---resulting in a time complexity of $t_{\Delta}=2T\cdot t_{\text{fwd}(\theta)}$. To alleviate this computational burden, we have proposed an approximation using a subset of timesteps $S$, which reduces $T$ to $|S|$. However, this approach also incurs additional time for sampling objectives to identify an optimal subset $S$. The time complexity of this approximation scheme is given by:
\begin{equation*}
    t_{\tilde{\Delta}} = 2 |S| \cdot t_{\text{fwd}(\theta)} + t_{S},
    \quad \text{where} \quad t_{S} = 2\frac{T}{|B|} \cdot t_{\text{fwd}(\theta)}.
\end{equation*}
The reason behind the time $t_{S}$ spent on subset selection is that we use a single $x_0$ to compute objective samples for this selection, while $t_{\text{fwd}}$ involves a full batch of $|B|$ $x_0$s during the forward pass. The time spent for running the feature selection algorithm is negligible (takes about $1\%$ of $t_{\text{fwd}}$) and is therefore ignored. With typical hyperparameter choices in practice, such as $|S|=3$, $|B|=128$, $T=1000$, this results in  $t_{\tilde{\Delta}}\approx 21\cdot t_{\text{fwd}(\theta)}$ for running \cref{alg:reward approximation}.

Since we run \cref{alg:reward approximation} only once every $f_S = 40$ updates of $\theta$, the total time for the proposed algorithm is given by:
\begin{align*}
    t_{\text{iter}(\theta, \phi)}&=t_{\text{iter}(\theta)} + t_{\text{fwd}(\phi)}+\frac{1}{f_S} \left(t_{\tilde{\Delta}}+t_{\text{iter}(\phi)}\right)\\
    &\approx 5.63\cdot t_{\text{fwd}(\theta)}\approx 1.41 \cdot t_{\text{iter}(\theta)},
\end{align*}
where we have assumed $t_{\text{fwd}(\theta)} \approx t_{\text{fwd}(\phi)}$. In practice, there are additional overheads due to minor factors, and we observed that our algorithm takes approximately 1.5 times longer in terms of wall-clock time compared to the baseline.

\section{Implementation Details}
\label{sec:implementation_details}
We mainly used a machine equipped with four NVIDIA RTX 3090 GPUs to train the models.

\begin{table}[ht!]
    \centering
    \begin{tabular}{lccc}
         Dataset &CIFAR-10 32$\times$32 &CelebA-HQ 256$\times$256 &ImageNet 256$\times$256  \\
         \toprule
         Diffusion architecture &DDPM~\citep{ho2020denoising} &LDM~\citep{rombach2022high} &ADM~\citep{dhariwal2021diffusion}
    \end{tabular}
    \caption{Our implementation details based on CIFAR-10, CelebA-HQ, ImageNet datasets.}
\end{table}

\begin{table}[ht!]
    \centering
    \begin{tabular}{lccc}
         Dataset &CIFAR-10 32$\times$32 &CelebA-HQ 256$\times$256 &ImageNet 256$\times$256  \\
         \toprule
         sampling steps &1000 &200 &50 \\
         sampling algorithm &DDPM sampler &DDIM sampler~\citep{song2020denoising} 
         &EDM sampler~\citep{karras2022elucidating} \\
         number of samples in evaluation &50K &50K &50K         
    \end{tabular}
    \caption{Our evaluation settings based on CIFAR-10, CelebA-HQ, ImageNet datasets.}
\end{table}

\paragraph{Baseline}
For all baselines, we used the most popular settings. For Min-SNR~\cite{hang2023efficient}, we used snr gamma=5. For P2~\cite{choi2022perception}, we used gamma=0 and k=1. For log normal~\cite{karras2022elucidating}, we sampled weights from a normal distribution with a mean of 0 and a standard deviation of 1, followed by applying a sigmoid function. For SpeeD~\cite{wang2024closer}, we sampled timesteps according to the official code and used gamma=1 and k=1 for weighting.

\paragraph{Feature Selection method}
To overcome the computational burden of calculating $\delta_{k,i}^t$ across all timesteps, we employ a feature selection method. Specifically, we treat $\delta^t_{k,i}$ as features to predict the target $\Delta_k^t$ using a linear regression model. The F-statistic is then computed for each feature $\delta^t_{k,i}$ and by summing the top $M$ features with the highest F-statistics, we can efficiently approximate $\Delta_k^t$ while focusing on the most influential timesteps $i$. The F-statistic identifies the features that have the strongest linear relationship with the target variable, enabling us to focus on the most critical timesteps in the approximation process. Although the F-statistic is employed here, other feature selection methods could also be applied.

\section{Hyperparameter settings}
\label{sec:hyperparameters}

\begin{table}[ht!]
    \centering
    \caption{Hyperparameter settings for CIFAR-10}
    \begin{tabular}{llr}
        \toprule
        \textbf{Category} & \textbf{Parameter} & \textbf{CIFAR-10} \\
        \midrule
        \multirow{20}{*}{\textbf{Diffusion}} 
         & Timesteps & 1000 \\
         & Beta Start & 0.0001 \\
         & Beta End & 0.02 \\
         & Beta Schedule & Linear \\
         & Model Mean Type & Eps \\
         & Model Variance Type & Fixed-large \\
         & Loss Type & MSE \\
         & Backbone & UNet \\
         & In Channels & 3 \\
         & Hidden Channels & 128 \\
         & Channel Multipliers & [1, 2, 2, 2] \\
         & Number of Residual Blocks & 2 \\
         & Drop Rate & 0.1 \\
         & Learning Rate & 2e-4 \\
         & Batch Size & 128 \\
         & Gradient Norm & 1.0 \\
         & Epochs & 2040 \\
         & Warmup & 5000 \\
         & Use EMA & True \\
         & EMA Decay & 0.9999 \\
        \midrule
        \multirow{7}{*}{\textbf{Timestep Sampler}} 
         & Learning Rate & 1e-2(linear, quad), 1e-3(cosine) \\
         & Entropy Coefficient & 1e-2 \\
         & In Channels & 3 \\
         & Hidden Channels & 128 \\
         & Hidden Depth  & 2 \\
         & $f_s$ & 40 \\
         & $|Q|$ & 20 \\
         & $|S|$ & 3 \\
        \bottomrule
    \end{tabular}
\end{table}

\newpage
\begin{table}[ht!]
    \centering
    \caption{Hyperparameter settings for CelebA-HQ 256x256}
    \begin{tabular}{llr}
        \toprule
        \textbf{Category} & \textbf{Parameter} & \textbf{CelebA-HQ}\\
        \midrule
        \multirow{16}{*}{\textbf{Diffusion}}
         & Timesteps & 1000  \\
         & Beta Schedule & Linear\\
         & Model Mean Type & Eps \\
         & Loss Type & MSE \\
         & Backbone & UNet \\
         & In Channels & 3 \\
         & Hidden Channels & 224 \\
         & Channel Multipliers & [1, 2, 3, 4] \\
         & Number of Residual Blocks & 2 \\
         & Drop Rate & 0.0 \\
         & Optimizer & AdamW $(\beta_1 = 0.9, \beta_2 = 0.999)$ \\
         & Learning Rate & 1.92e-4  \\
         & Batch Size & 24  \\
         & Epochs & 300  \\
         & Use EMA & True \\
         & EMA Decay & 0.9999 \\
        \midrule
        \multirow{7}{*}{\textbf{Timestep Sampler}}
         & Learning Rate & 1e-3  \\
         & Entropy Coefficient & 1e-2 \\
         & In Channels & 3 \\
         & Hidden Channels & 256 \\
         & Hidden Depth  & 2 \\
         & $f_s$ & 40 \\
         & $|Q|$ & 20 \\
         & $|S|$ & 3 \\
        \bottomrule
    \end{tabular}
\end{table}

\newpage
\begin{table}[ht!]
    \centering
    \caption{Hyperparameter settings for ImageNet 256x256}
    \begin{tabular}{llr}
        \toprule
        \textbf{Category} & \textbf{Parameter} & \textbf{ImageNet}\\
        \midrule
        \multirow{15}{*}{\textbf{Diffusion}} 
         & Timesteps & 1000  \\
         & Beta Schedule & Cosine\\
         & Model Mean Type & Epsilon \\
         & Loss Type & MSE \\
         & Backbone &U-ViT \\
         & Layers & 238  \\
         & Hidden Size & 1152  \\
         & Heads & 16 \\
         & Depths & 12 \\
         & Optimizer & AdamW $(\beta_1 = 0.99, \beta_2 = 0.99)$ \\
         & Learning Rate & 2e-4  \\
         & Batch Size & 256  \\
         & Training iterations & 2.1M  \\
         & Use EMA & True \\
         & EMA Decay & 0.9999  \\
        \midrule
        \multirow{7}{*}{\textbf{Timestep Sampler}} 
         & Learning Rate & 1e-3  \\
         & Entropy Coefficient & 1e-2 \\
         & In Channels & 3 \\
         & Hidden Channels & 256 \\
         & Hidden Depth  & 2 \\
         & $f_s$ & 40 \\
         & $|Q|$ & 20 \\
         & $|S|$ & 3 \\
        \bottomrule
    \end{tabular}
\end{table}

\newpage
\section{Visualization}
\label{sec:Visualization}
\begin{figure*}[h!]
    \centering
    \includegraphics[width=1.0\linewidth]{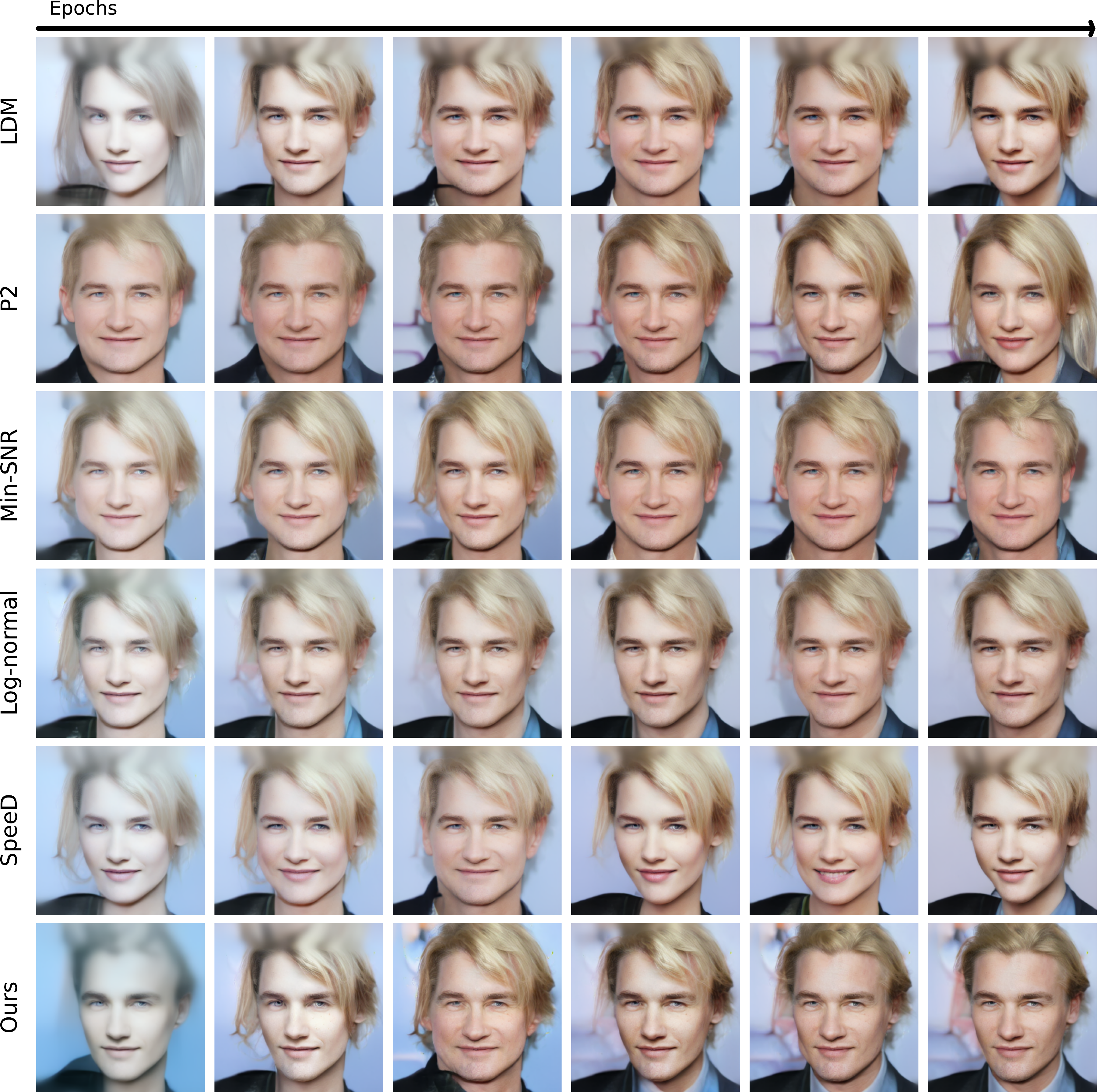}
    \caption{Visualization of CelebA-HQ images generated by our method and baseline methods.}
    \label{fig:viz_celebahq}
\end{figure*}

\end{document}